\documentclass{article}
\usepackage[final,nonatbib]{neurips_2019}

\usepackage{amssymb}
\usepackage{amsmath}
\usepackage{enumerate}
\usepackage[utf8]{inputenc} 
\usepackage[T1]{fontenc}   
\usepackage[colorlinks=true]{hyperref}     
\usepackage{url}  
\usepackage[dvipsnames]{xcolor}
\usepackage{booktabs}       
\usepackage{amsfonts}      
\usepackage{nicefrac}    
\usepackage{textcomp}    
\usepackage{gensymb}
\usepackage{graphicx}

\title{Scaling active inference}
\author{
  Alexander Tschantz$^{1,2}$ \\
  \texttt{tschantz.alec@gmail.com} \\
  \And
  Manuel Baltieri$^{2,3}$ \\ 
  \And
  Anil. K. Seth$^{1,4}$ \\
  \And
  Christopher L. Buckley$^{2}$ \\
  \\
  $^{1}$Sackler Centre for Consciousness Science, University of Sussex, Brighton, UK \\
  $^{2}$Evolutionary and Adaptive Systems Research Group, University of Sussex, Brighton, UK \\
  $^{3}$RIKEN Centre for Brain Science, Saitama, Japan \\
  $^{4}$Canadian Institute for Advanced Research, University of Sussex, Brighton, UK \\ \\
}

\begin{document}
\maketitle

\begin{abstract}
  In reinforcement learning (RL), agents often operate in partially observed and uncertain environments.
  Model-based RL suggests that this is best achieved by learning and exploiting a probabilistic model of the world.
  ‘Active inference’ is an emerging normative framework in cognitive and computational neuroscience that offers a unifying account of how biological agents achieve this.
  On this framework, inference, learning and action emerge from a single imperative to maximize the Bayesian evidence for a niched model of the world.
  However, implementations of this process have thus far been restricted to low-dimensional and idealized situations.
  Here, we present a working implementation of active inference that applies to high-dimensional tasks, with proof-of-principle results demonstrating efficient exploration and an order of magnitude increase in sample efficiency over strong model-free baselines.
  Our results demonstrate the feasibility of applying active inference at scale and highlight the operational homologies between active inference and current model-based approaches to RL.
\end{abstract}


\section{Introduction}
\label{sec:intro}

In model-based reinforcement learning (RL), agents first learn a predictive model of the world, before using this model to determine actions \cite{atkeson_comparison_1997}.
Encoding a model of the world plausibly affords several advantages.
For instance, such models can be used to perform perceptual inference \cite{ha_recurrent_2018}, implement prospective control \cite{chua_deep_2018,schrittwieser_mastering_2019}, quantify and resolve uncertainty \cite{shyam_model-based_2019}, and generalize existing knowledge to new tasks and environments \cite{hafner_learning_2018}. 
As such, the use of predictive models has been touted as a potential solution to the sample inefficiencies of modern RL algorithms \cite{deisenroth_pilco:_2011, schmidhuber_making_1990}.

At the same time, the theoretical framework of active inference has emerged in cognitive and computational neuroscience as a unifying account of perception, action, and learning \cite{friston_free-energy_2010, friston_active_2017}. 
Active inference suggests that biological systems learn a probabilistic model of their habitable environment and that the states of the system change to maximize the evidence for this model \cite{friston_active_2015,friston_free_2019}. 
The resulting scheme casts perception, action and learning as emergent processes of (approximate) Bayesian inference, thereby offering a potentially unifying theory of adaptive biological systems.
Despite its strong theoretical foundations, existing computational implementations have been restricted to low-dimensional tasks, often with discrete state spaces and actions \cite{friston_active_2015,friston_active_2017,friston_active_2017-1,friston_active_2017-2,friston_deep_2018}.
Here, we establish a formal connection between active inference and model-based RL. 
In doing so, we extend practical implementations of active inference so that they work effectively at scale, and we situate model-based RL within the broad theoretical context offered by active inference. 

We present a model of active inference that is applicable in high-dimensional control tasks with both continuous states and actions. 
Our model builds upon previous attempts to scale active inference \cite{millidge_deep_2019,ueltzhoffer_deep_2018,catal_bayesian_2019} by including an efficient planning algorithm, as well as the quantification and active resolution of model uncertainty.
Consistent with the active inference framework, learning and inference are achieved by maximizing single lower bound on Bayesian model evidence, and policies are selected to maximize a lower bound on expected Bayesian model evidence \cite{friston_active_2015}.
We demonstrate that this unified normative scheme enables sample efficient learning, strong performance on difficult control tasks, and a principled approach to active exploration.
Moreover, we establish homologies between our active inference based model and state-of-the-art approaches to model-based RL.

In what follows, we specify the general mathematical formulation of active inference, before describing our implementation, which is applicable in both partially-observed and fully-observed environments.
We then present preliminary results in three challenging fully-observed continuous control benchmarks, leaving the analysis of partially-observed environments (i.e. pixels) to future work. 
These results demonstrate that our algorithm facilitates active exploration over long temporal horizons and significantly outperforms a strong model-free RL baseline, in terms of both sample efficiency and performance.


\section{Active inference}
\label{sec:active inference}

Following previous work \cite{friston_active_2017,friston_active_2015}, we consider active inference in the context of a partially observed Markov decision process (POMPD). 
At each time step $t$, the true state of the environment $\mathbf{\hat{s}}_t \in \mathbb{R}^{d_{\hat{s}}}$ evolves according to the stochastic transition dynamics $\mathbf{\hat{s}_t} \sim \mathrm{p}(\mathbf{\hat{s}}_t|\mathbf{\hat{s}}_{t-1}, \mathbf{a}_{t-1})$, where $\mathbf{a} \in \mathbb{R}^{d_{a}}$ denotes an agent's actions. 
Agents do not always have access to the true state of the environment, but might instead receive observations $\mathbf{o}_t \in \mathbb{R}^{d_{o}}$, which are generated according to $\mathbf{o}_t \sim\textbf{}\mathrm{p}(\mathbf{o}_t|\mathbf{\hat{s}}_t)$. 
As such, agents must operate on \textit{beliefs} $\mathbf{s}_t \in \mathbb{R}^{d_{s}}$ about the true state of the environment $\mathbf{\hat{s}}_t$. 
In what follows, we denote the true dynamics with upright letters $\mathrm{p}(\cdot)$ and a model of these dynamics (the agent) with italics $p(\cdot)$. 

Active inference proposes that agents implement and update a generative model of their world $p(\tilde{\mathbf{o}}, \tilde{\mathbf{s}}, \pi, \theta)$, where the tilde notation denotes a sequence of variables through time $\tilde{\mathbf{x}} = \{ \mathbf{x}_0 ,..., \mathbf{x}_T \}$, $\pi$ denotes a \textit{policy}, $\pi = \{\mathbf{a}_0, ..., \mathbf{a}_T\}$, and $\theta \in \Theta$ denotes parameters of the generative model, which are themselves random variables. 
Additionally, agents maintain a recognition distribution $q(\tilde{\mathbf{s}},\pi, \theta)$, representing an agent's (approximately optimal) beliefs over states $\tilde{\mathbf{s}}$, policies $\pi$ and model parameters $\theta$. 

As new observations are sampled, agents update the parameters of their recognition distribution to minimize variational \textit{free energy} $\mathcal{F}$: 

\begin{equation}
\begin{aligned}
\label{eq:free-energy}
\mathcal{F}(\tilde{\mathbf{o}}) &= \mathbb{E}_{q(\tilde{\mathbf{s}},\pi, \theta)}[\ln q(\tilde{\mathbf{s}},\pi, \theta) - \ln p(\tilde{\mathbf{o}}, \tilde{\mathbf{s}}, \pi, \theta)] \\
&\geq -\ln p(\tilde{\mathbf{o}})
\end{aligned}
\end{equation}

This makes the recognition distribution $q(\tilde{\mathbf{s}},\pi, \theta)$ converge towards an approximation of the (intractable) posterior distribution $p(\tilde{\mathbf{s}},\pi, \theta|\tilde{\mathbf{o}})$, thereby implementing a tractable form of (approximate) Bayesian inference \cite{blei_variational_2017}.

Crucially, active inference also proposes that an agent's goals and desires are encoded in the generative model as prior preferences for favourable observations \cite{friston_free_2019, baltieri2019pid}, i.e. blood temperature at 37\degree C. 
Free energy then provides a proxy for how surprising (i.e., unlikely) some observations are under the agent's model. 
While minimising Eq. \ref{eq:free-energy} provides an estimate for how surprising some observations are, it cannot reduce this quantity directly. 
To achieve this, agents must change their observations through action. 
Acting to minimise variational free energy ensures the minimisation of \emph{surprisal} $-\ln p(\tilde{\mathbf{o}})$, or the maximisation of the (Bayesian) \emph{model evidence} $p(\tilde{\mathbf{o}})$, since free energy provides an upper bound on surprisal.
Active inference, therefore, proposes that agent's select policies in order to minimize \textit{expected} free energy $\mathcal{G}$ \cite{friston_free_2019}, where the expected free energy for a given policy $\pi$ at some future time $\tau$ is:

\begin{equation}
\begin{aligned}
\label{eq:expected-free-energy}
\mathcal{G}(\pi, \tau) &= \mathbb{E}_{q(\mathbf{o}_{\tau},\mathbf{s}_{\tau}, \theta|\pi)}[\ln q(\mathbf{s}_{\tau}, \theta | \pi) - \ln p(\mathbf{o}_{\tau}, \mathbf{s}_{\tau}, \theta| \pi)] \\
&\geq -\mathbb{E}_{q(\mathbf{o}_\tau|\pi)}\big[\ln p(\mathbf{o}_{\tau}|\pi)\big]
\end{aligned}
\end{equation}

Expected free energy provides a bound on \textit{expected} surprisal, and can be decomposed into \textit{extrinsic} value, which quantifies the degree to which expected observations are congruent with an agent's prior beliefs, and \textit{intrinsic} value, which quantifies the amount of information an agent expects to gain from enacting some policy \cite{friston_active_2017-1,friston_active_2015, friston_active_2017}. 
This decomposition affords a natural interpretation: to avoid being surprised, one should sample unsurprising data, but also learn about the world to make data less surprising \textit{per se}. 
Selecting policies that minimize Eq. \ref{eq:expected-free-energy} will, therefore, ensure that probable (i.e. favourable, given an agent's normative priors) observations are preferentially sampled, while also ensuring that agents gather information about their environment.


\section{Model}
\label{sec:model}

In cognitive and computational neuroscience, implementations of active inference agents generally follow one of two approaches. The first considers the generative model and recognition distribution to be Gaussian under the Laplace approximation and prescribes gradient-descent updates that recurrently minimize free energy with each new observation \cite{friston_predictive_2009, buckley_free_2017, baltieri2019pid}.
While this approach is purported as biologically plausible and enjoys empirical support under the guise of predictive coding \cite{friston_predictive_2009, clark_whatever_2013}, it is not clear how, or at least not straightforward, to extend this implementation to the prospective free energy minimization discussed in Sec. \ref{sec:active inference}.
The second approach employs discrete distributions (e.g., Categorical, Dirichlet) that are updated via variational message-passing \cite{friston_active_2015}. 
While this approach provides an elegant framework for evaluating expected free energy, it can only be applied in discrete state and action spaces, meaning it is not directly applicable to the high-dimensional states and continuous actions considered in RL benchmarks.

In the current paper, we take an alternative approach and employ \textit{amortized} inference \cite{kingma_auto-encoding_2013}, which utilizes function approximators (i.e., neural networks) to parameterize distributions. 
Free energy is then minimized with respect to the parameters of the function approximators, and not the variational parameters themselves. 
We detail our generative model and recognition distribution in Sec. \ref{sec:gen-model}, how learning and inference are implemented in Sec. \ref{sec:learning}, how policy selection and trajectory sampling are implemented in Sec. \ref{sec:policy} \& Sec. \ref{sec:trajectory}, and how to evaluate expected free energy in section Sec. \ref{sec:expected}. 
Finally, we describe the implementation details for the fully-observed case in Sec. \ref{sec:implementation-details}.


\subsection{Generative model \& recognition distribution}
\label{sec:gen-model}

We consider a generative model $p(\tilde{\mathbf{o}}, \tilde{\mathbf{s}}, \pi, \theta)$ over sequences of observations $\tilde{\mathbf{o}}$, hidden states $\tilde{\mathbf{s}}$, policies $\pi$ and parameters $\theta$:

\begin{equation}
\label{eq:generative-model}
\begin{aligned}
    p(\tilde{\mathbf{o}}, \tilde{\mathbf{s}}, \pi, \theta) &= p(\theta) p(\pi) \prod^{T}_{t=1}p(\mathbf{o}_t|\mathbf{s_t}) p(\mathbf{s_t}|\mathbf{s}_{t-1},\pi_{t-1}, \theta) \\
    p(\mathbf{o}_t|\mathbf{s_t}) &= \mathcal{N}(\mathbf{o}_t;\mu_{\lambda}, \sigma^2_{\lambda}) \\ 
    & \mathrm{where} \ [\mu_{\lambda}, \sigma^2_{\lambda}] = f_{\lambda}(\mathbf{s}_t) \\ 
    p(\mathbf{s_t}|\mathbf{s}_{t-1},\pi_{t-1}, \theta) &= \mathcal{N}(\mathbf{s}_t; \mu_{\theta}, \sigma^2_{\theta}) \\ 
    & \mathrm{where} \ [\mathrm{\mu_{\theta}}, \sigma^2_{\theta}] = f_{\theta}(\mathbf{s}_{t-1},\pi_{t-1}) \\
    p(\theta) &=  \mathcal{N}(\theta; 0,\mathbb{I}) \\ 
    p(\pi) &= \sigma(-\mathcal{G}(\pi)) 
\end{aligned}
\end{equation}

where we have assumed that ${s}_{0}$ is fixed. 
In Eq. \ref{eq:generative-model}, we have parametrized both the likelihood distribution $p(\mathbf{o}_t|\mathbf{s_t})$ and the transition distribution $p(\mathbf{s_t}|\mathbf{s}_{t-1},\pi_{t-1}, \theta)$ with function approximators. 
Specifically, the likelihood distribution is described by a multivariate Gaussian distribution with a mean and covariance parameterized by some (potentially non-linear) function approximator $f_{\lambda}(\mathbf{s}_t)$, while the prior distribution is described by a Gaussian with mean and variance parameterized by some function approximator $f_{\theta}(\mathbf{s}_{t-1},\pi_{t-1})$. 

Amortizing the inference procedure offers several benefits. 
For instance, the number of parameters remains constant with respect to the size of the data and inference can be achieved through a single forward pass of a network. 
Moreover, while the amount of information encoded about variables is fixed, the conditional relationship between variables can be arbitrarily complex. 
In Eq. \ref{eq:generative-model}, the parameters of the transition distribution, $\theta$, are themselves random variables. 
In the current context, these parameters are the weights of the neural network $f_{\theta}(\mathbf{s}_{t-1},\pi_{t-1})$. 
This approach allows the uncertainty about these parameters to be quantified and casts learning as a process of (variational) inference \cite{blundell_weight_2015}. 
The prior probability of $\theta$ is given by a standard Gaussian, which acts as a regularizer during learning. Although we have only considered distributions over the parameters of the transition distribution $\theta$, the same scheme could be applied to the parameters of the likelihood distribution, $\lambda$. 
Finally, the prior probability of policies is a softmax function of the negative expected free energy of those policies $-\mathcal{G}(\pi)$ \cite{friston_active_2015}. 
This formalizes the notion that policies are \textit{a-priori} more likely if they are expected to minimize free energy in the future \cite{friston_free_2019}.

To make active inference applicable to the kinds of tasks considered in RL, we treat reward signals $\mathbf{o}^r$ as observations in a separate modality. 
Therefore, we extend the generative model to include an additional scalar Gaussian over reward observations $p(\mathbf{o}^r_t|\mathbf{s}_t)$ with unit variance and mean $f_{\alpha}(\mathbf{s}_t)$, where $f_{\alpha}(\mathbf{s}_t)$ is a fully-connected neural network with parameters $\alpha$.

We consider a recognition distribution $q(\tilde{\mathbf{s}}, \pi, \theta)$ over sequences of hidden states $\mathbf{s}_t$, policies $\pi$ and parameters $\theta$:

\begin{equation}
\begin{aligned}
\label{eq:recognition-distribution}
q(\tilde{\mathbf{s}}, \pi, \theta) &= q(\theta)q(\pi)\prod^T_{t=0}q(\mathbf{s}_t|\mathbf{o}_t) \\
q(\theta) &= \mathcal{N}(\theta; \mu_{\xi}, \sigma^2_{\xi}) \\
q(\pi) &= \mathcal{N}(\pi; \mu_{\psi}, \sigma^2_{\psi}) \\
q(\mathbf{s_t}|\mathbf{o}_t) &= \mathcal{N}(\mathbf{s_t}; \mu_{\phi}, \sigma^2_{\phi}) \\
& \mathrm{where} \ [\mu_{\phi}, \sigma^2_{\phi}] = f_{\phi}(\mathbf{o}_t) \\ 
\end{aligned}
\end{equation}

The distribution $q(\mathbf{s_t}|\mathbf{o}_t)$ is a diagonal Gaussian with mean and variance parameterized by some function approximator $f_{\phi}(\mathbf{o}_t)$, while the the variational posterior over parameters $\theta$ and policies $\pi$ are both diagonal Gaussians.


\subsection{Learning \& Inference}
\label{sec:learning}

In order to implement learning, we derive the updates for $\xi = \{\mu_{\xi}, \sigma^2_{\xi}\}$, $\phi$, $\lambda$ and $\alpha$ that minimize free energy $\mathcal{F}$. 
Given Eq. \ref{eq:generative-model} and \ref{eq:recognition-distribution}, the variational free energy $\mathcal{F}$ for a given time point $t$ can be defined as:

\begin{equation}
\begin{aligned}
\label{eq:free-energy-params}
  &\mathcal{F}_t(\mathbf{o}_t, \xi, \phi, \lambda, \alpha) = \\
  &\mathbb{E}_{\theta \sim q(\theta)}\Big[\mathbb{E}_{ q(\mathbf{s}_{t-1}|\mathbf{o}_{t-1})}\big[\mathbf{D}_{KL}[q(\mathbf{s}_t|\mathbf{o}_t)||p(\mathbf{s}_t|\mathbf{s}_{t-1},\pi_{t-1},\theta)]\big]\Big] \\
&\ \ \ \ \ \ \ \ \ \ \ \ \ + \mathbf{D}_{KL}\big[q(\theta)||p(\theta)\big] - \mathbb{E}_{q(\mathbf{s}_{t}|\mathbf{o}_{t})}[\ln p(\mathbf{o}_t|\mathbf{s}_t)]
\end{aligned}
\end{equation}

where we have followed \cite{friston_active_2015} and omitted the additional term $\mathbf{D}_{KL}[q(\pi)||p(\pi)]$ from the optimisation of $\xi, \phi, \lambda, \alpha$, allowing us to ignore the dependency between hidden states and (the prior probability of) policies. 
We optimise $q(\pi)$ with respect to $\mathcal{F}$ separately, as described in the following section.

Eq. \ref{eq:free-energy-params} can be minimized with respect to $\xi, \phi, \lambda, \alpha$ using stochastic gradient descent. 
Given some observation $\mathbf{o}_t$, the negative log-likelihood (third term) can be calculated by mapping the observation to the variational parameters of $q(\mathbf{s}_t|\mathbf{o}_t)$, e.g.,  $[\mu_{\phi},\sigma^2_{\phi}] = f_{\phi}(\mathbf{o}_t)$. 
The reparameterization trick \cite{kingma_auto-encoding_2013} is then utilized to obtain a differentiable sample from $q(\mathbf{s}_t|\mathbf{o}_t)$\footnote{For a Gaussian $\mathcal{N}(\mathbf{x}; \mu, \sigma^2)$, a reparameterized sample is obtained via $\mathbf{x} = \mu+  \sigma \odot \epsilon$, where $\epsilon \sim \mathcal{N}(0, 1)$}, which is then passed through $f_{\lambda}(\mathbf{s}_t)$, giving the parameters of the likelihood distribution $[\mu_{\lambda}, \sigma_{\lambda}^2]$. 
The negative-log likelihood of the observations is then calculated under this distribution. 
Next, the parameter divergence (second term) is calculated analytically, as both distributions are fully factorized Gaussians. 
Finally, The state divergence (first term) is calculated by taking $K$ samples from $q(\theta)$, again using the reparameterization trick. 
For each sample $\theta^{(i)}$ in $K$, a reparameterized sample from the previous beliefs over hidden states $q(\mathbf{s}_{t-1}|\mathbf{o}_{t-1})$ is propagated through $f_{\theta^{(i)}}(\mathbf{s}_{t-1}, \pi_{t-1})$ (where $\pi_{t-1}$ refers to the action that was taken at the previous time step), giving the parameters of the transition distribution. 
The KL-divergence term is then analytically calculated for each sample in $K$ and averaged.

This procedure is carried out in batched fashion over the available data set. 
At test time, inference can be achieved by directly mapping observations to the variational parameters using $f_{\phi}(\mathbf{o}_t)$. 
This approach to inferring hidden states is similar to that of a variational autoencoder \cite{kingma_auto-encoding_2013}, but here the global prior has been replaced with a prior based on the transition distribution. 
Moreover, the inference of parameters $\theta$ is homologous to the Bayesian neural network approach to parameter learning \cite{blundell_weight_2015}. 

Deriving updates for all parameters through a single (variational) objective function offers several potential benefits. 
First, the learned latent space is forced to balance between the compression of observations and (action-conditioned) temporal transitions.
This is in contrast to `modular' approaches, whereby a latent space is first learned to compress observations, and subsequently, a transition model is learned in this fixed latent space \cite{ha_recurrent_2018}.
Moreover, this approach allows the quantification of uncertainty in both hidden states \textit{and} model parameters, thereby quantifying both aleatoric and epistemic uncertainty \cite{depeweg_decomposition_2017,depeweg_decomposition_2017-1}.


\subsection{Policy selection}
\label{sec:policy}

Under active inference, policy selection is achieved by updating $q(\pi)$ in order to minimize free energy $\mathcal{F}$. 
Given the prior belief that policies minimize expected free energy, i.e., $p(\pi) = \sigma(-\mathcal{G}(\pi))$ (as specified in Eq. \ref{eq:generative-model}), free energy is minimized when $q(\pi) = \sigma(-\mathcal{G}(\pi))$ \cite{friston_active_2015}. 
For discrete action spaces with short temporal horizons, $\mathcal{G}(\pi)$ can be evaluated in full by considering each possible policy \cite{friston_active_2017}. 
However, in continuous action spaces, there are infinite policies, meaning an alternative approach is required. 

In the current work, we treat $q(\pi)$ as a diagonal Gaussian with parameters $\psi = \{ \mu_{\psi}, \sigma^2_\psi \}$. At each time step, we optimise $\psi$ such that $q(\pi) \propto -\mathcal{G}(\pi)$. 
While this solution will fail to capture the exact shape of $-\mathcal{G}(\pi)$, agents need only identify the peak of the landscape to enact the optimal policy. 
To optimise the parameters of $q(\pi)$, we utilise the cross-entropy method (CEM) \cite{hafner_learning_2018,chua_deep_2018}. 
At each time step $t$, we consider policies of a fixed horizon $H$, using notation $\pi^{t:t+H} = \{\mathbf{a}_t , ... , \mathbf{a}_{t+H} \}$. 
The distribution over policies is initialized as $q(\pi^{t:t+H}) \leftarrow \mathcal{N}(\pi^{t:t+H}; 0, \mathbb{I})$ and optimized as follows:
\begin{itemize}
\item[\textbf{(i)}] Sample $N$ policies from $q(\pi^{t:t+H})$
\item[\textbf{(ii)}] Evaluate $-\mathcal{G}(\pi^{t:t+H})$ for each sample $\pi^{t:t+H}$ (described in the following section), returning a scalar value
\item[\textbf{(iii)}] Refit $q(\pi^{t:t+H})$ to the top $M$ samples
\end{itemize}

This procedure is carried out $I$ times, after which the mean of the belief for the current time step $\mathbf{a}_t = \mathbb{E}[q(\pi^{t:t+H}_t)]$ is returned.
Moreover, this procedure is carried out after each new observation.
For the current experiments, $H = 12$, $N = 1000$, $M = 100$ and $I = 10$.

This process of model predictive control \cite{camacho_model_2007} was selected for consistency with previous computational models of active inference \cite{friston_active_2017}, where a distribution over policies is updated after each new observation. 
Alternative approaches include optimizing a parametrized policy with respect to past evaluations of expected free energy \cite{millidge_deep_2019}. 
However, this approach is not suited for non-stationary objective functions or active exploration \cite{shyam_model-based_2019}. 
Alternatively, a parametrized policy could be optimized with respect to imagined rollouts from a transition model \cite{hafner_learning_2018}, which \textit{would} enable active exploration \cite{shyam_model-based_2019}. 
The effectiveness of these approaches depends on the complexity of the value function relative to the transition dynamics \cite{dong_bootstrapping_2019}, as well as the stationarity of the value function.


\subsection{Trajectory sampling}
\label{sec:trajectory}

To evaluate the expected free energy for a given policy $-\mathcal{G}(\pi)$, it is first necessary to evaluate the expected future beliefs conditioned on that policy $q(\tilde{\mathbf{s}}^{t:t+H}, \tilde{\mathbf{o}}^{t:t+H}|\pi)$. 
The fact that the transition model is probabilistic, and the parameters of the transition model are random variables, induces a distribution over future trajectories \cite{friston_active_2015}. 
Several approaches exist to approximate the propagation of uncertain trajectories \cite{chua_deep_2018}. 
For instance, one can ignore uncertainty entirely and propagate the mean of the distributions, or one can explicitly propagate the full statistics of the distribution \cite{deisenroth_gaussian_2015}. 
In the current work, we utilise a \textit{particle} approach \cite{chua_deep_2018,hafner_learning_2018}, whereby a set of Monte Carlo samples are propagated. 
In particular, we consider $B$ samples from the parameter distribution $\theta^{(i)} \sim q(\theta)$, and for each sample in $B$, propagate $J$ samples through the transition model $\mathbf{s}_{t}^{(j)} \sim p(\mathbf{s}_t|\mathbf{s}_{t-1}, \pi_{t-1}, \theta^{(i)})$. 
To infer observations and rewards, we pass all samples through the respective model and average.


\subsection{Expected free energy}
\label{sec:expected}

In this section we describe how to evaluate $-\mathcal{G}(\pi)$, where we have used $\pi = \pi^{t:t+H}$ for notational convenience. 
The negative expected free energy for a policy is equal to the sum of negative expected free energies over time, $-\mathcal{G}(\pi) = \sum_{\tau=t}^{t+H}-\mathcal{G}(\pi, \tau)$, where 

\begin{equation}
\begin{aligned}
\label{eq:expected-free-energy-param}
-\mathcal{G}(\pi, \tau) &\approx \underbrace{\mathbb{E}_{q(\mathbf{o}^r_{\tau}|\pi)}[\ln p(\mathbf{o}_{\tau}^r)]}_\text{Extrinsic value} \\
&+ \underbrace{\mathbf{H}[q(\mathbf{o}_{\tau}|\pi)] - \mathbb{E}_{q(\mathbf{s}_{\tau}|\pi)}\Big[\mathbf{H}[q(\mathbf{o}_{\tau}|\mathbf{s}_{\tau}, \pi)]\Big]}_\text{State information gain}  \\ 
&+ \underbrace{\mathbf{H}[q(\mathbf{s}_{\tau}|\pi)] - \mathbb{E}_{q(\theta)}\Big[\mathbf{H}[q(\mathbf{s_{\tau}}|\pi, \theta)]\Big]}_\text{Parameter information gain}
\end{aligned}	
\end{equation}

We refer to \cite{friston_active_2017} for a derivation of Eq. \ref{eq:expected-free-energy-param}. 
The first term (\textit{extrinsic value}) quantifies the degree to which the expected observations $q(\mathbf{o}_{\tau}^r|\pi)$ are congruent with the agent's prior beliefs (i.e., preferences) $p(\mathbf{o}^r_t)$. 
Note that in active inference, there is no intrinsic delineation of reward signals - all observations are assigned some \textit{a-priori} probability. 
However, as RL environments specify a distinct reward signal, we have defined the agent's prior preferences over reward observations $\mathbf{o}^r$ only. 
Moreover, as RL environments are constructed such that agents wish to simply \textit{maximize} the sum of rewards (rather than obtain any particular reward observation), we evaluate extrinsic value as $\mathbf{o}_\tau^r \sim q(\mathbf{o}_{\tau}|\pi)$, such that extrinsic value increases as larger rewards are predicted. 
We refer the reader to \cite{catal_bayesian_2019} for an alternative formulation where agent's \textit{learn} a specific prior distribution. 

The second term (\textit{state information gain}) quantifies the expected reduction in uncertainty in beliefs over hidden states $q(\mathbf{s}_{\tau})$. 
In other words, it promotes agents to sample data in order to resolve uncertainty about the hidden state of the environment. 
This term is formally equivalent to a number of established quantities, such as (expected) Bayesian surprise, mutual information, and the expected reduction in posterior entropy \cite{friston_active_2015,tschantz_learning_2019}, and has been used to describe various epistemic foraging behaviors, such as saccades \cite{parr_discrete_2018,yang_active_nodate,itti_bayesian_2009,mirza_introducing_2019} and sentence comprehension \cite{friston_deep_2018}. 
In the current paper, we conduct experiments in fully observed environments, and as such, do not consider the state information gain term in our analysis. 

The final term (\textit{parameter epistemic value}) quantifies the expected reduction in uncertainty in beliefs over parameters $q(\theta)$, and promotes agents to actively explore the environment in order to resolve uncertainty in their model \cite{schwartenbeck_computational_2019, friston_active_2017-2}. 
In order to evaluate parameter epistemic value, we use a nearest neighbor estimate of the entropies \cite{depeweg_decomposition_2017-1,beirlant_nonparametric_1997}. 
In other words, we estimate the entropy via spatial properties of samples from the relevant distributions.
Specifically, we estimate the entropy as $\mathbf{H}[p(\mathbf{x})] = \frac{1}{n} \sum_{i=1}^{n} \ln \left(n \rho_{n, i}\right)+\ln 2+C_{E}$, where $n$ is the number of samples from the distribution, $\rho_{n, i}$ is the nearest neighbor distance of a sample $\mathbf{x}_i$ from other samples $\mathbf{x}_j$ and $C_{E}$ is the Euler constant. 
Alternatively, parameter epistemic value could be rewritten as the (expected) Bayesian surprise of the distribution over parameters and then calculated analytically \cite{houthooft_vime:_2016, barron_information_2018,itti_bayesian_2009,mirza_introducing_2019}. 
However, this requires doing fictive updates to the parameter distribution, which is computationally expensive when conducted for each candidate policy at each time step.


\subsection{Fully observed model}
\label{sec:implementation-details}

The model presented in the preceding sections serves as the most general formulation, applicable in both partially-observed and fully-observed environments.
In what follows, we describe an implementation for the fully-observed case, leaving an analysis of the partially-observed case for future work.

To adapt the generative model for fully-observed environments, we utilise a fixed identity covariance for the likelihood distribution $p(\mathbf{o}_t|\mathbf{s}_t)$, and parameterize the mean as $\mu_{\lambda} = f_{\lambda}(\mathbf{s}_t) = \mathbb{E}[\mathbf{s}_t]$, thereby encoding the belief that there is a direct mapping between states and observations. 
For the transition distribution $p(\mathbf{s_t}|\mathbf{s}_{t-1},\pi_{t-1}, \theta)$, we parameterize the mean as $f_{\theta}(\mathbf{s}_{t-1},\pi_{t-1})$ and utilize a fixed unit variance. 
In all experiments, $f_{\theta}(\mathbf{s}_{t-1},\pi_{t-1})$ is a feed-forward network with two fully connected layers of size 500 with ReLU activations, which defines the dimensionality of $p(\theta)$ and $q(\theta)$.

Note that by treating the variance of the transition distribution as fixed, the evaluation of parameter epistemic value is significantly simplified. 
Specifically, the second entropy term in parameter epistemic value becomes constant under policies, such that we need only evaluate the first entropy term $\mathbf{H}[q(\mathbf{s}_{\tau}|\pi)] = \mathbf{H}[\mathbb{E}_{q(\theta)}[q(\mathbf{s}_{\tau}|\pi, \theta)]]$.
We use 5 samples from $q(\theta)$ to evaluate the expectation in this entropy term throughout. 
Finally, we treat the variance of $q(\mathbf{s}_t|\mathbf{o}_t)$ as a fixed unit parameter and parameterize the mean as $\mu_{\phi} = f_{\phi}(\mathbf{o}_t) = \mathbf{o}_t$, thereby encoding the belief that there is a direct mapping between observations and states. 
Note that this means that the parameters of $\lambda$ and $\phi$ are fixed and are thus excluded from the optimisation scheme.


\section{Experiments}
\label{sec:experiments}

In this section, we investigate (i) whether the proposed active inference model can successfully promote exploration in the absence of reward observations (i.e. exploration), and (ii) whether the model can achieve good performance and high sample efficiency on challenging continuous control tasks (i.e. exploitation). 
We evaluate these two aspects of the model separately, leaving analysis of their joint performance (i.e. the exploration-exploitation dilemma) to future work.

We utilise the following learning scheme for both the exploration and exploitation experiments. We initialize the data set with 5 seed episodes collected under random actions. 
For each iteration of the experiment, we train the agent's model via Eq. \ref{eq:free-energy-params} with 100 batches randomly sampled from the data set, using a batch size of 50. 
Agents then collect data from the environment until the episode ends (when the maximum number of steps is reached, or when agent the agent reaches a terminal state).


\subsection{Exploration}

\begin{figure}[htb]
\centering
\centerline{\includegraphics[width=12.0cm]{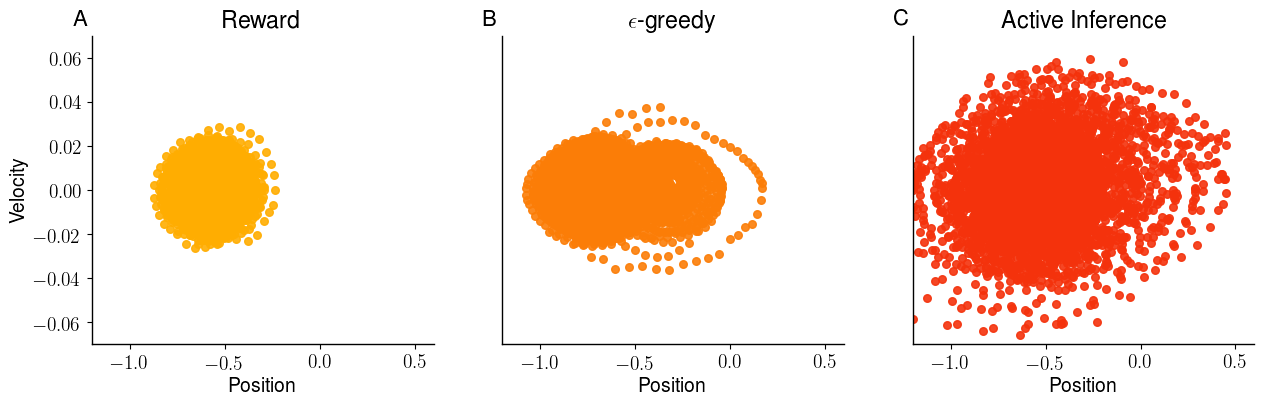}}
\caption{\textbf{Comparison of exploration strategies}. (A) The cumulative state-space coverage after 100 epochs for the reward agent. (B) The cumulative state-space coverage after 100 epochs for the $\epsilon$-greedy agent. (C) The cumulative state-space coverage after 100 epochs for the active inference agent. These results demonstrate that the active inference agent explores more of the state space, relative to the other exploration strategies.}
\label{fig:1}
\end{figure}

To test whether the active inference model enables efficient exploration, we explore the state space visited by different algorithms in the continuous MountainCar environment ($\mathcal{O} \in \mathbb{R}^2, \mathcal{A} \in \mathbb{R}^1$). 
We compare the active inference model to two algorithms, (i) a `reward' agent which operates via the same scheme, but only selects actions based on extrinsic value, and (ii) and an $\epsilon$-greedy agent which selects action based on extrinsic value, but additionally adds Gaussian exploration noise ($\sigma^2 = 0.3$) to each action. 
Agents learn and act in the environment for 100 epochs. The cumulative coverage of state space is plotted in Fig. \ref{fig:1}. 
These results demonstrate that the active inference agent can effectively explore more of the state space, relative to the other algorithms.

\subsection{Exploitation}

Next, we investigate whether the active inference agent can achieve good performance on continuous control tasks. We explore performance in the inverted pendulum task ($\mathcal{O} \in \mathbb{R}^3$, $\mathcal{A} \in \mathbb{R}^1$) and the more challenging hopper task ($\mathcal{S} \in \mathbb{R}^{15}$, $\mathcal{A} \in \mathbb{R}^3$). 
The performance of our model is compared to a strong model-free baseline, DDPG \cite{lillicrap_continuous_2019}. 

As both environments have well-shaped rewards, we only consider the exploitation component (extrinsic value) of the expected free energy objective function, ignoring the exploration component (epistemic value). 
As such, we utilise a point-estimate version of the model, thus removing the distributions over parameters. 
To retain stochasticity in the transition model, we parameterize both the mean and variance of the transition distribution using a function approximator (as opposed to just the mean), and fix the variance of the recognition distribution to 0.1. 
Moreover, following \cite{hafner_learning_2018}, we use an action repeat of 3 for all algorithms, enabling shorter planning horizons and a more pronounced learning signal. 

\begin{figure}[htb]
  \centering
  \centerline{\includegraphics[width=12.0cm]{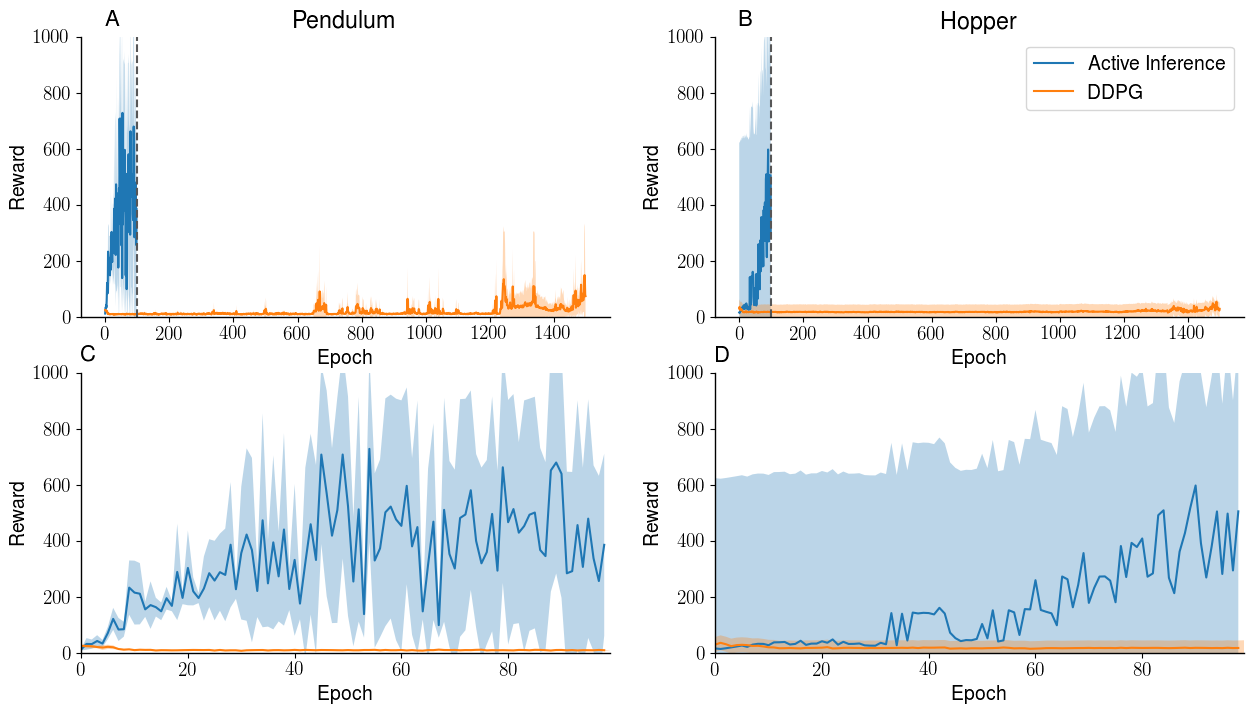}}
  \caption{\textbf{Comparison of Performance on two continuous control tasks}. (A) Average returns over 1500 epochs on the inverted pendulum task for the active inference agent and the model-free DDPG agent. (B) Average returns over 1500 epochs on the hopper task for the active inference agent and the model-free DDPG agent. Note that for A \& B, we stopped the active inference agent after 100 epochs due to convergence. (C \& D) A zoomed-in view of figures A \& B, showing a more fine-grained view of the active inference agent's progression over 100 epochs. For all figures, the filled lines represent the mean of 5 random seeds, whereas the shaded areas denote the 95\% interquartile range. Together, these results demonstrate that the active inference agent can learn difficult continuous control tasks, with a far greater sample efficiency, relative to a strong model-free baseline.}
  \label{fig:2}
\end{figure}

In Fig. \ref{fig:2}, we plot the performance of both algorithms as a function of the number of epochs. 
We show the mean performance over a fixed set of 5 random seeds and the shaded lines shown the 95\% interquartile range at each epoch. 
These results demonstrate that the active inference agent can achieve strong performance in under 100 epochs on both tasks, demonstrating an order of magnitude better sample efficiency compared to the model-free baseline.

\section{Previous work}
\label{sec:previous work}

\paragraph{Deep active inference} Previous work has explored the prospect of scaling active inference using amortized inference. 
In \cite{ueltzhoffer_deep_2018}, the authors parameterized both the generative model and recognition distribution with function approximators and used evolutionary strategies to optimise the free energy functional when gradients were not available. 
Similarly, \cite{millidge_deep_2019} utilized amortization to parametrize distributions and also amortized action by learning a parameterized approximation of the expected free energy bound. 
Finally, \cite{catal_bayesian_2019} extended previous work to include a specific planning component based on CEM. The authors focused on the problem of learning the prior distribution over reward observations $p(\mathbf{o}^r)$ and demonstrated this could be implemented in a learning-from-example framework.

Our work builds upon these previous models by incorporating model uncertainty and its active resolution. 
In other words, we extend the previous point-estimate models to include full distributions over parameters and update the expected free energy functional such that the uncertainty in these distributions is actively minimized. 
This brings our implementation in line with the canonical models of active inference from the cognitive and computational neuroscience literature \cite{friston_free_2019}. 
Moreover, it enables us to evaluate the feasibility of active exploration under the scaled active inference framework, apply the model to more complex control tasks, and obtain increased sample efficiency, relative to previous models.


\paragraph{Model-based RL} The model presented in the current work bears a number of resemblances with model-based approaches to RL. 
First, variational autoencoders have been used extensively to map observations into a compressed latent space, thereby simplifying the problem of action selection and the process of learning a forward transition model \cite{ha_recurrent_2018,hafner_learning_2018,igl_deep_2018, karl_deep_2016,kaiser_model-based_2019,barron_information_2018,watter_embed_2015}. 
Moreover, the CEM algorithm is a popular method for implementing planning in model-based RL \cite{hafner_learning_2018,chua_deep_2018, nagabandi_neural_2017}. 
Recent work has additionally highlighted the importance of using a probabilistic dynamics model in order to capture epistemic uncertainty \cite{chua_deep_2018,hafner_learning_2018,deisenroth_pilco:_2011, yarin_gal_improving_2016, kahn_uncertainty-aware_2017,vuong_uncertainty-aware_2019}. The success of these approaches has demonstrated that deterministic models are prone to model bias, which can lead to overfitting in low data regimes. 
Most approaches either utilize Bayesian neural networks \cite{depeweg_decomposition_2017-1}, ensembles of deterministic networks \cite{chua_deep_2018}, dropout \cite{yarin_gal_improving_2016} or Gaussian processes \cite{deisenroth_gaussian_2015} in order to capture uncertainty. 
In the current work, we opted for Bayesian neural networks to ensure consistency with the variational principles espoused by the active inference framework, but note that ensembles can be made explicitly Bayesian with minor modifications \cite{pearce_bayesian_2018}.


\paragraph{Information gain} Identifying scalable and efficient exploration strategies remains one of the key open questions in RL. 
Model-free methods, such as $\epsilon$-greedy or Boltzmann choice rules \cite{sutton_introduction_1998}, utilise noise in the action selection process or uncertainty in the reward statistics \cite{agrawal_analysis_2012, speekenbrink_uncertainty_2015}. 

A more powerful approach \cite{osband_generalization_2016} is to construct a model of the world, allowing the agent to evaluate which parts of state space it has (and has not) visited. 
For instance, \cite{bellemare_unifying_2016} construct a pseudo-count measure for estimating state visitation frequency in continuous state spaces. 
Alternatively, an explicit forward model of the transition dynamics can be learned. 
This allows for measures such as the amount of prediction error \cite{stadie_incentivizing_2015, thrun_efficient_1992, chentanez_intrinsically_2005, meyer_possibility_1991} or prediction error improvement \cite{lopes_exploration_2012} to be utilized for exploration.

If the learned model (implicitly or explicitly) captures probabilistic features then information-theoretic measures can be used to guide exploration (see \cite{aubret_survey_2019} for a review). 
In \cite{still_information-theoretic_2012}, the authors derived an information-theoretic measure to maximize the predictive power of the agent, while in \cite{mohamed_variational_2015}, the authors derived an objective function to maximize the mutual information between actions and future states of the environment (i.e., empowerment).

Of particular relevance to the current work is the use of \textit{information gain} to promote exploration, which has been demonstrated to outperform alternative measures such as prediction error \cite{hester_intrinsically_2017}.
From a theoretical perspective, information gain helps overcome what is known as the "TV-problem" \cite{itti_bayesian_2009}, where (unpredictable) noise in the environment is mistakenly treated as epistemically valuable.
This is because information gain considers the amount of information provided for \textit{beliefs}, as opposed to the amount of information provided by the signal \textit{per se}. 

Information gain can be traced back to \cite{lindley_measure_1956}, who used it to quantify the amount of information to be gained from some experiment.
\cite{sun_planning_2011} developed a Bayesian framework in order to maximize information gain via dynamic programming, however, the experiments were limited to discrete state spaces using tabular MDPs. In \cite{houthooft_vime:_2016}, the authors utilized Bayesian neural networks to quantify the amount of information gained from some (action-conditioned) transition.
This work was further extended in \cite{barron_information_2018}, where the amount of information gain was quantified with respect to a latent dynamics model. 

In parallel with the current work, \cite{shyam_model-based_2019} looked to maximize \textit{expected} information gain, which entails an \textit{active} approach to exploration.
This is in contrast to the majority of exploration strategies in RL, which are \textit{reactive}, in the sense that they must first observe an informative state before being able to gather information \cite{shyam_model-based_2019}. 
This can lead to problems of over-commitment, whereby informative parts of state space must be unlearned once the relevant information has been gathered. 
However, \cite{shyam_model-based_2019} optimized expected information gain offline, whereas the current model uses an online approach. 
Finally, The use of nearest-neighbour entropy estimators for information gain has been explored in \cite{mirchev_approximate_2018, depeweg_decomposition_2017-1}.


\section{Discussion \& Conclusion}

We have presented a model of active inference that can scale to continuous control tasks, complex dynamics and high-dimensional state spaces. 
The presented model can be trained via a single objective function, expected free energy, that captures both epistemic and aleatoric uncertainty, and prescribes both goal-directed and information-gathering behaviour via a single normative drive.

Our model makes two primary contributions. 
First, we showed that the full active inference construct can be scaled to the kinds of tasks considered in the RL literature. 
This involved extending previous models of deep active inference to include model uncertainty and expected information gain. 
Second, we highlighted the overlap between active inference and state-of-the-art approaches to model-based RL. 
These include the use of variational inference for the compression of observations, the use of variational inference for learning distributions over parameters, the use of probabilistic models of dynamics, the use of prospective planning in latent space, and the active resolution of uncertainty. 

While active inference defined the properties of living systems from first principles \cite{friston_free_2019}, and model-based RL has attempted to engineer adaptive agents through the most effective means available, both perspectives have converged on similar solutions. 
Our work has exploited this convergence to show that active inference provides a principled and unified theoretical framework in which to contextualize the various developments in model-based RL. 
This perspective by itself offers little practical benefit. 
However, active inference offers two potentially novel perspectives from which model-based RL can benefit. The first is casting reward as (prior) probabilities. 
This provides a principled framework for learning reward structure (i.e., reward shaping), for assigning rewards (i.e., probability) across multiple observation modalities \cite{juechems_where_2019}, and for learning-from-demonstration \cite{catal_bayesian_2019}.
The second perspective is casting both exploration and exploitation as two components of a single imperative to maximize expected Bayesian model evidence. 
This perspective has the potential to recast the exploration-exploitation dilemma as a problem of optimising parameters in order to maximise model evidence. 
We leave a practical investigation of this perspective to future work.


\section{Acknowledgements}

AT is funded by a PhD studentship from the Dr. Mortimer and Theresa Sackler Foundation and the School of Engineering and Informatics at the University of Sussex. 
CLB is supported by BBRSC grant number BB/P022197/1. MB acknowledges support as an International Research Fellow of the Japan Society for the Promotion of Science. 
AT and AKS are grateful to the Dr. Mortimer and Theresa Sackler Foundation, which supports the Sackler Centre for Consciousness Science. 
AKS is additionally grateful to the Canadian Institute for Advanced Research (Azrieli Programme on Brain, Mind, and Consciousness). 

\bibliographystyle{unsrt}

\end{document}